\documentclass{article} % ICLR 2027 (variant of the NeurIPS format)
\PassOptionsToPackage{nonamebreak}{natbib}
\usepackage{iclr2027_conference,times}
\ifdefined\pdfminorversion\pdfminorversion=6\relax\fi

\usepackage{amsmath,amsfonts,bm}

\def\eqref#1{equation~\ref{#1}}

\def\1{\bm{1}}

\DeclareMathAlphabet{\mathsfit}{\encodingdefault}{\sfdefault}{m}{sl}
\SetMathAlphabet{\mathsfit}{bold}{\encodingdefault}{\sfdefault}{bx}{n}

\usepackage{hyperref}
\hypersetup{
  hidelinks,
  pdftitle={ScaleNav: Scalable Internet Video Pretraining for Language-Conditioned Local Navigation},
  pdfauthor={Yihui Yang, Yuanhong Yu, Jiawei Wang, Zhongrui Yu, Jia Shi, Kun Zhan, Ruizhen Hu, Xiaowei Zhou, Yu Liu, Sida Peng}
}
\usepackage{xurl}
\usepackage{graphicx}
\usepackage{amsmath,amssymb}
\usepackage{booktabs}
\usepackage{colortbl,xcolor}
\usepackage{multirow}
\usepackage{placeins}
\AddToHook{env/table/begin}{\setlength{\belowcaptionskip}{\baselineskip}}
\usepackage{fvextra}
\fvset{fontsize=\normalsize,breaklines=true,breaksymbolleft={},frame=single,framesep=2mm,framerule=0.3pt}

\definecolor{lightpink}{RGB}{233, 23, 115}
\definecolor{first}{RGB}{220, 245, 220}   % softer light green for 1st place
\definecolor{second}{RGB}{255, 255, 200}  % light yellow for 2nd place
\definecolor{third}{RGB}{200, 230, 255}   % light blue for 3rd place

\title{ScaleNav: Scalable Internet Video \\
Pretraining for Language-Conditioned \\
Local Navigation}

\author{
Yihui Yang\textsuperscript{1}\quad Yuanhong Yu\textsuperscript{1}\quad Jiawei Wang\textsuperscript{1}\quad Zhongrui Yu\textsuperscript{2}\quad Jia Shi\textsuperscript{2}\\[0.4em]
\bfseries Kun Zhan\textsuperscript{2}\quad Ruizhen Hu\textsuperscript{3}\quad Xiaowei Zhou\textsuperscript{1}\quad Yu Liu\textsuperscript{2}\quad Sida Peng\textsuperscript{1}\\[0.7em]
\normalfont\textsuperscript{1}Zhejiang University\quad \textsuperscript{2}Li Auto Inc.\quad \textsuperscript{3}Shenzhen University
}

\iclrfinalcopy

\begin{document}
\raggedbottom

\maketitle
\lhead{Preprint}
\suppressfloats[t]

\begin{abstract}
We study language-conditioned local navigation, where a robot must navigate to a text-specified target visible in its initial observation and stop at the described location. End-to-end navigation policies require diverse training data to generalize across scenes, yet real-robot data is costly to collect and simulated data leaves a sim-to-real gap. Internet egocentric videos offer a scalable source of real-world navigation experience. Our key insight is to learn goal-directed visual dynamics from Internet egocentric videos, and use these dynamics to support robot action generation. To this end, we propose ScaleNav, a unified world-action framework with three-stage training. We first perform instruction-conditioned video pretraining on these videos to learn how observations evolve as the camera approaches instructed targets across diverse real-world scenes. Action alignment then uses action-labeled simulated data to train the model to predict executable action chunks. Joint fine-tuning subsequently optimizes all trainable components together for navigation. ScaleNav achieves success rates of 86.5\% on a single-view adaptation of Short-Horizon OVON and 94.9\% on our HabitatGS-LN benchmark, outperforming the evaluated baselines on both benchmarks. We further deploy and evaluate ScaleNav on a real quadruped robot, with results supporting the model's ability to generalize to real-world navigation. Code and videos are available at \url{https://zju3dv.github.io/ScaleNav/}.
\end{abstract}

\section{Introduction}
\label{sec:intro}

\begin{figure}[t]
\centering
\includegraphics[width=\linewidth]{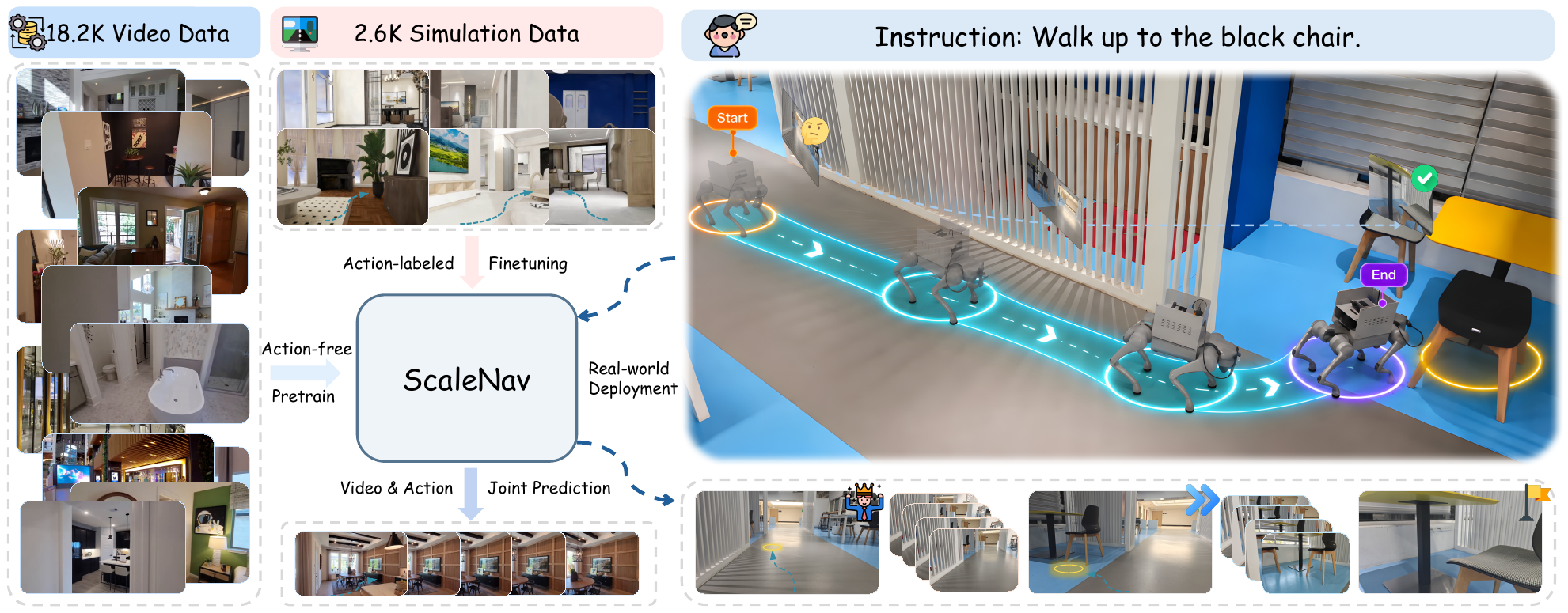}
\caption{\textbf{Overview of ScaleNav.} ScaleNav learns navigation priors from Internet egocentric videos through action-free video pretraining, then uses action-labeled simulated data for action alignment and joint fine-tuning. It jointly predicts future observations and action chunks to follow navigation instructions. The right panel illustrates deployment on a real quadruped robot.}
\label{fig:teaser}
\end{figure}

Navigating to visible targets according to natural-language instructions is an important capability for robots. For example, a person may ask a robot to walk to the desk by the window and stop there. Similar instructions also arise in long-horizon systems, where a high-level planner, such as an agent based on a large language model (LLM), decomposes a task into a sequence of short-range goals involving visible objects or regions~\citep{huang2023vlmaps,rajvanshi2024saynav,ding2026lavira}. We study this language-conditioned local navigation setting: the robot must navigate to a text-specified target visible in its initial field of view and stop at the described location.

One possible solution is to connect a grounding module to an existing navigation method~\citep{shah2022lmnav,huang2023vlmaps,cai2024pixnav,xu2025mobilityvla,long2024instructnav,padillacerdio2026openfrontier}, but errors in target grounding and spatial-goal conversion can lead to navigation errors. End-to-end policies provide an alternative by mapping visual observations and language instructions directly to actions~\citep{zhang2024navid,hirose2024lelan,uninavid2025,wei2026streamvln,zhang2025navfom,zhu2026navida,wang2026lightnav0,huang2026wnm3d}. To generalize across real-world scenes, these policies require diverse training data~\citep{shah2023gnm,shah2023vint,sridhar2024nomad}. However, real-robot action data is costly to collect across diverse environments~\citep{liu2025citywalker,navdp2026}, whereas simulated action data still leaves a sim-to-real gap~\citep{navdp2026,xia2026habitatgs}.

Internet egocentric videos offer a scalable source of visual experience covering diverse real-world scenes and human navigation behavior. However, they lack corresponding robot action labels. Some methods attempt to leverage these videos by extracting intermediate motion-related supervision, such as estimated trajectories and geometric targets~\citep{cheng2025navila,liu2025citywalker,hirose2024lelan,seneviratne2026vega,zhou2026uninav}, or latent actions inferred from visual transitions~\citep{bu2025univla}. Errors or biases introduced during this extraction can propagate through the resulting supervision into subsequent policy learning.

\begin{figure}[!t]
\centering
\includegraphics[width=\linewidth]{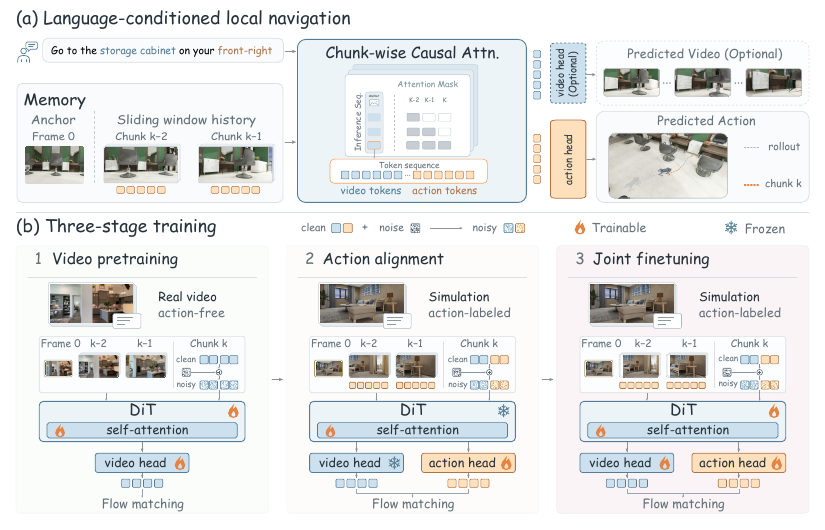}
\caption{\textbf{Overview of ScaleNav. (a) Language-conditioned local navigation.} The model predicts video and action chunks from the instruction, current observation, and memory using chunk-wise causal attention. The memory contains an initial-frame anchor and a sliding window of recent video and action chunks. \textbf{(b) Three-stage training.} Video pretraining uses action-free real videos, followed by action alignment and joint fine-tuning with action-labeled simulation data. During action alignment, the self-attention layers within the DiT, the action head, and the action embedding MLP (not shown) are trainable. The snowflake on the outer DiT block marks the remaining DiT layers as frozen. Joint fine-tuning updates the DiT and both heads. For clarity, the initial bidirectional adaptation in Stage 1 and Internet egocentric video replay in Stages 2 and 3 are omitted.}
\label{fig:method-pipeline-draft}
\end{figure}

To effectively learn navigation priors from these videos and translate them into executable actions, we propose ScaleNav, a framework for language-conditioned local navigation that combines a unified world-action model (WAM) with a three-stage training procedure. As illustrated in Figure~\ref{fig:method-pipeline-draft}, the model uses a shared Diffusion Transformer (DiT)~\citep{peebles2023dit} to jointly predict future observations and action chunks. We first perform instruction-conditioned video pretraining on Internet egocentric videos to learn navigation priors from human navigation behavior in diverse real-world scenes. Action alignment then uses action-labeled simulated data to train the model to predict executable action chunks while keeping part of the model frozen. Finally, joint fine-tuning optimizes all trainable components together for video and action prediction in navigation. During navigation, the model predicts an action chunk from the instruction and visual observations, executes actions from the chunk, and repeats this process with updated observations until it reaches the target and stops.

For evaluation, we construct HabitatGS-LN, a benchmark for navigation to specific objects described in natural-language instructions. We also evaluate ScaleNav on a single-view adaptation of Short-Horizon OVON~\citep{gong2026abotn1}. ScaleNav achieves success rates (SR) of 86.5\% on Short-Horizon OVON and 94.9\% on HabitatGS-LN, outperforming the compared methods on both benchmarks. With identical action-labeled simulated data, incorporating Internet egocentric videos improves success by 5.3 and 4.5 percentage points on the two benchmarks, respectively. We further deploy and evaluate ScaleNav on a real quadruped robot in three indoor environments. ScaleNav achieves 83.3\% overall SR and outperforms the evaluated baselines in all three environments.

Our contributions are as follows:
\begin{itemize}

\item We propose ScaleNav, a framework for language-conditioned local navigation that transfers navigation priors learned from diverse Internet egocentric videos through video pretraining into executable action chunks via action alignment.
\item Experiments show that ScaleNav outperforms the compared methods in both navigation success rate and path efficiency on the two simulation benchmarks. Real-robot experiments further support ScaleNav's ability to generalize to real-world navigation.
\end{itemize}

\section{Related Work}
\label{sec:related}

\paragraph{Language-conditioned navigation.}
Language-conditioned navigation connects language understanding with visual perception, planning, and action generation.
Modular approaches combine pretrained foundation models with explicit maps, candidate waypoints, or planning components~\citep{shah2022lmnav,huang2023vlmaps,chen2024mapgpt,long2024instructnav,padillacerdio2026openfrontier}.
Learned vision-language-action policies adapt pretrained vision-language backbones to predict navigation actions from instructions and visual observations~\citep{zhang2024navid,uninavid2025,wei2026streamvln,zhang2025navfom,zhu2026navida,wang2026lightnav0}.
Within this direction, dual-system architectures connect high-level semantic reasoning to low-level action generation~\citep{wei2025dualvln,gong2026abotn1}.
Within language-conditioned navigation, local navigation focuses on reaching a language-specified destination without a prescribed route.
Methods for this setting learn language-conditioned goal-reaching policies~\citep{hirose2024lelan,glossop2025cast,mei2026urbannav} or address local target grounding under occlusion~\citep{gao2026beacon}.
We focus on this local navigation setting.

\paragraph{World-action models.}
Using video prediction for robot control requires linking visual predictions to executable actions~\citep{zhao2026worldvln}.
One approach generates visual plans and recovers actions through inverse dynamics or geometric estimation~\citep{du2023unipi,ko2024avdc,chen2026imaginav,lee2026cuenav}.
Related approaches condition their action decoders on predictive visual features or latents~\citep{hu2025vpp,zhang2026sparsevideonav}.
Another line of work jointly models future observations and actions within a unified architecture~\citep{guo2024pad,zhu2025uwm,cen2025worldvla,ye2026dreamzero,dong2026uniwm,azuma2026navwam,yang2026wamnav,zhou2026uninav,huang2026wnm3d,li2026metis}.
Following DreamZero, we adopt this joint video-action architecture for navigation.

\paragraph{Learning from Internet egocentric videos.}
Prior work in robot manipulation learns from action-free human videos through video pretraining~\citep{wu2024gr1,cheang2024gr2} or video-only cross-embodiment transfer~\citep{ye2026dreamzero}.
In navigation, existing approaches leverage these videos mainly in two ways.
One line of work uses Internet videos to train navigation policies through intermediate motion-related supervision or latent-action representations.
Geometry-based methods derive trajectory or geometric prediction targets from estimated camera motion or reconstructed scenes~\citep{liu2025citywalker,cheng2025navila,mei2026urbannav,han2025roomtour3d,seneviratne2026vega,zhou2026uninav}.
Other methods obtain action annotations or teacher supervision from learned models~\citep{lin2023youtubevln,hirose2024lelan}.
Related video-based methods infer latent actions from visual transitions for policy pretraining~\citep{ye2025lapa,bu2025univla}.
Errors or biases in these derived targets and representations can propagate into subsequent policy learning.

Another line of work uses Internet videos to learn world models for navigation planning.
NWM learns a navigation world model from unlabeled videos without proxy motion targets~\citep{bar2025nwm}.
\citet{garrido2026latentworld} learn a latent-action world model from Internet videos and adapt it to navigation by mapping robot actions to latent actions.
These models evaluate candidate trajectories determined by a separate search procedure or policy.
Better predictions therefore do not necessarily yield better navigation decisions when suitable candidates are missing.
Our framework learns navigation priors from Internet egocentric videos and directly predicts executable action chunks, without constructing proxy motion targets for these videos or relying on an external planner at inference.

\section{Method}
\label{sec:method}

\subsection*{Overview}

We study language-conditioned local navigation: given a text instruction and a stream of RGB observations, a robot navigates to a target object or region visible in its initial field of view and stops at the instructed location. A single action $a_t=(\Delta x_t,\Delta y_t,\Delta\psi_t)$ represents the motion between two consecutive frames, where $(\Delta x_t,\Delta y_t)$ is the planar displacement in the robot's local coordinates at the preceding frame and $\Delta\psi_t$ is the change in yaw. A sequence of $T$ actions forms an action chunk, denoted by $\tau=(a_1,\ldots,a_T)$. The robot collects RGB observations while executing each predicted action chunk, then uses the updated observations to predict the next chunk.

We use a world-action model with a shared DiT backbone to jointly predict future observations and action chunks (Section~\ref{sec:arch}). Through three-stage training (Section~\ref{sec:training}), the model first learns navigation priors from Internet egocentric videos and is then adapted for action prediction using action-labeled simulated data. Section~\ref{sec:data} describes how we construct both datasets.

\subsection{Model Architecture}
\label{sec:arch}

\paragraph{Unified world-action model.}We follow DreamZero~\citep{ye2026dreamzero} and use a unified world-action model that jointly predicts future observations and actions with a shared DiT backbone. Because the two modalities share backbone parameters, action prediction can draw on visual priors learned from Internet egocentric videos. Specifically, we denote the clean variational autoencoder (VAE)~\citep{kingma2014vae} latents of future observations by $z_0^v$ and the clean, normalized representation of action chunk $\tau$ by $z_0^a$. Their noisy counterparts at noise levels $t_v$ and $t_a$ are denoted by $z_{t_v}^v$ and $z_{t_a}^a$, respectively. The noisy video and action representations are embedded as tokens, concatenated, and jointly processed by the DiT, conditioned on the navigation instruction $l$, current observation $o$, and history. The video and action heads map the corresponding output tokens to flow velocities $v_\theta^v$ and $v_\theta^a$ for joint denoising, where $\theta$ denotes the model parameters. The denoised action representation is denormalized into an action chunk, and the video latents are optionally decoded by the VAE into future observations.

\paragraph{Initial-frame anchor and recent history.}Task instructions are specified with respect to the initial observation, and the robot must use the visual context in that view to identify the intended navigation target. For example, in ``walk over to the second door on your left'', ``second'' refers to the ordering of doors in the initial view. To maintain accurate instruction interpretation as the viewpoint changes, we use an anchor mechanism that retains the initial frame's video tokens throughout navigation. We also maintain a sliding window containing multiple recent chunks, each consisting of observations and past actions. The initial-frame anchor and the recent chunks form the history condition $M$ to support coherent navigation across chunks. This history is passed to the DiT as noise-free tokens, which current-chunk tokens attend to through chunk-wise causal attention.

\subsection{Three-Stage Training}
\label{sec:training}

We use a three-stage training procedure to learn navigation priors from Internet egocentric videos and use them to support action chunk generation. Stage 1 pretrains the model to predict egocentric observations as the camera approaches and reaches the instructed target, conditioned on navigation instructions. Stage 2 introduces action-labeled simulated data for action alignment, learning action prediction with part of the model frozen. Stage 3 jointly fine-tunes the model for video and action prediction.

\paragraph{Stage 1: Video pretraining.}
Pretrained video models provide spatiotemporal priors for action generation~\citep{ye2026dreamzero}. However, their text conditioning, training-data selection, and prediction schemes differ from the requirements of downstream navigation. First, their text conditions typically describe scene content and motion~\citep{kong2024hunyuanvideo,yang2024cogvideox}, whereas navigation instructions specify a target to reach. Second, during data preparation, camera-dominated footage with little subject motion and clips with pronounced camera shake are sometimes downweighted or excluded~\citep{wan2025wan,yang2024cogvideox}, limiting exposure to viewpoint changes and camera shake, both of which are common during robot navigation. Finally, these models commonly generate entire clips bidirectionally~\citep{yang2024cogvideox,yin2025causvid}, whereas navigation requires updating predictions as new observations arrive.

To bridge these differences and further learn navigation priors from human navigation in diverse real-world scenes, we continue pretraining Wan2.2-TI2V-5B~\citep{wan2025wan} on our Internet egocentric videos paired with navigation instructions. To avoid changing the data distribution and prediction structure simultaneously, we first train on these video-instruction pairs using Wan's original bidirectional text-image-to-video (TI2V) formulation. Throughout Stage 1, we condition on the navigation instruction and current observation, keeping the observation latent noise-free. We then continue training with chunk-wise prediction, using the initial-frame anchor and recent video chunks as noise-free history, corresponding to the video component of $M$. Chunk-wise causal attention allows current-chunk tokens to attend to historical video tokens while retaining bidirectional attention within the current chunk.

Throughout video pretraining, we use a video flow-matching objective~\citep{lipman2023flowmatching} with no action tokens in the input. For the future video latents $z_0^v$, we sample a noise level $t_v\in(0,1)$ and Gaussian noise $\epsilon^v\sim\mathcal{N}(0,I)$, and construct $z_{t_v}^v=(1-t_v)z_0^v+t_v\epsilon^v$. The model predicts the corresponding flow velocity $v_\theta^v$ with the objective
\[
\mathcal{L}_{\mathrm{video}}=\mathbb{E}\left[\left\|v_\theta^v-(\epsilon^v-z_0^v)\right\|^2\right].
\]
The VAE and text encoder remain frozen throughout all three training stages.

\paragraph{Stage 2: Action alignment.}To translate the navigation priors learned during video pretraining into executable actions, we train the model to jointly predict future observations and action chunks using action-labeled simulated data. During this alignment stage, we only train the randomly initialized action embedding multilayer perceptron (MLP) and action head together with the shared self-attention layers.

We independently sample video and action noise levels $t_v,t_a\in(0,1)$~\citep{zhu2025uwm} and add noise to the video latents as in Stage 1. For the normalized action chunk $z_0^a$, we sample Gaussian noise $\epsilon^a\sim\mathcal{N}(0,I)$ and construct $z_{t_a}^a=(1-t_a)z_0^a+t_a\epsilon^a$. The model jointly predicts the video and action flow velocities $v_\theta^v$ and $v_\theta^a$. For batches of action-labeled simulated data, we normalize the inverse-loss weights discussed by \citet{ma2024harmonydream} to sum to one, yielding the training objective
\[
\begin{aligned}
\mathcal{L}&=(1-\lambda)\mathcal{L}_{\mathrm{video}}+\lambda\mathcal{L}_{\mathrm{action}},\\
\mathcal{L}_{\mathrm{action}}&=\mathbb{E}\left[\left\|v_\theta^a-(\epsilon^a-z_0^a)\right\|^2\right],\\
\lambda&=\operatorname{sg}\!\left(\frac{\mathcal{L}_{\mathrm{video}}}{\mathcal{L}_{\mathrm{video}}+\mathcal{L}_{\mathrm{action}}}\right).
\end{aligned}
\]
Here, $\lambda$ is computed from the two losses on the current mini-batch, and $\operatorname{sg}$ denotes the stop-gradient operator. In Stages 2 and 3, we also replay Internet egocentric videos from the Stage 1 dataset to help retain the learned video priors, using only the video loss for these samples.

In addition, stopping after arrival is an important navigation capability. We model this behavior implicitly through action prediction. Specifically, we pad incomplete final chunks to the fixed length by repeating the last video frame and appending zero-motion actions $(0,0,0)$ before action normalization. To avoid potentially disturbing the pretrained video priors with artificially repeated final frames, we exclude latent positions marked as padding from $\mathcal{L}_{\mathrm{video}}$ by applying a temporally max-pooled frame-validity mask. The padded zero actions remain included in $\mathcal{L}_{\mathrm{action}}$ to supervise zero-motion outputs after arrival at the target.

\paragraph{Stage 3: Joint fine-tuning.}Action alignment trains the action modules while keeping much of the video-pretrained model frozen. To further adapt the shared representations to robot navigation together with the action modules, we jointly fine-tune the model. Specifically, we unfreeze the remaining DiT modules and the video head, using the same joint training objective as in Stage 2.

\subsection{Data}
\label{sec:data}

We construct two datasets for training: action-free Internet egocentric videos and action-labeled simulated data with aligned video and robot actions.

\paragraph{Internet egocentric videos.}We seek egocentric video clips that capture human navigation behavior across diverse real-world scenes. We use the scene annotations provided by SpatialVID-HQ~\citep{wang2025spatialvid} to select approximately 60K indoor clips from YouTube. For each clip, we provide uniformly sampled keyframes to a vision-language model (VLM) to identify candidate endpoints where the camera has arrived near a describable object or region. For each candidate endpoint, we take the corresponding object or region as the navigation target and use the VLM to search backward for the earliest keyframe in which the target is visible. Using the selected start and end timestamps, we extract the continuous segment from the source clip. For each extracted segment, we use the VLM to generate a navigation instruction describing the target by its category, attributes, or spatial relations.

To ensure accurate annotations and plausible navigation behavior, we then review the candidate segments and their instructions using the VLM. After filtering, the Internet egocentric video dataset contains 18.2K navigation segments. Detailed data construction and review procedures are provided in Appendix~\ref{app:data}.

\paragraph{Action-labeled simulated data.}To obtain navigation data with aligned video and actions, we generate episodes in Habitat-GS~\citep{xia2026habitatgs} using indoor scenes from InteriorGS~\citep{miao2026towards}. We select targets from the initial observations and use the VLM with scene semantic annotations to generate unambiguous navigation instructions. Using ground-truth navmeshes, we plan smooth, collision-free trajectories with an offline 2D adaptation of SCAN-Planner~\citep{zheng2026scanplanner} and collect aligned observations and action labels along these trajectories. We review the resulting samples using criteria similar to those for the Internet egocentric videos, with scene semantic annotations providing additional evidence for instruction accuracy. This yields approximately 2.6K action-labeled episodes from 650 scenes. Further construction details are provided in Appendix~\ref{app:data}.

For both datasets, the VLM paraphrases the instructions of accepted samples to increase language diversity while preserving their meaning.

\FloatBarrier
\section{Experiments}
\label{sec:exp}

\subsection{Simulation Experiments}
\label{sec:exp-setup}
\paragraph{Benchmarks.}
We first evaluate on Short-Horizon OVON~\citep{gong2026abotn1}, which repositions the starts of HM3D-OVON Val-Unseen episodes~\citep{yokoyama2024hm3dovon} to the first pose where the target becomes visible. The benchmark was originally developed for a three-view setting. We adapt it to our single-view setting, yielding 524 episodes across 36 HM3D scenes~\citep{ramakrishnan2021hm3d}, each paired with a navigation instruction referring to the target object.

Short-Horizon OVON evaluates navigation to category-level goals, but does not explicitly assess whether an agent can identify the intended object from the attributes and spatial relations described in an instruction. To complement this evaluation, we construct HabitatGS-LN, a benchmark for language-conditioned local navigation in photorealistic Habitat-GS environments~\citep{xia2026habitatgs}, with 157 episodes across 45 held-out InteriorGS scenes~\citep{miao2026towards}. Detailed adaptation and construction procedures are provided in Appendix~\ref{app:benchmark}.

\paragraph{Evaluation protocol.}
We conduct closed-loop evaluation on Short-Horizon OVON in Habitat~\citep{savva2019habitat} and on HabitatGS-LN in Habitat-GS~\citep{xia2026habitatgs}. Episodes terminate according to each method's stopping rule or after 200 execution steps, without using ground-truth target proximity to trigger stopping. Each executed action counts as one step, including each action within a predicted chunk. Success requires stopping within the budget and within 1.25~m of a valid target, measured by the minimum Euclidean distance between their footprints. Valid targets are annotated goal instances of the specified category on Short-Horizon OVON and the instruction-specified instance on HabitatGS-LN. We report success rate (SR), success weighted by path length (SPL)~\citep{anderson2018evaluation}, and amortized inference latency per executed action on RTX 4090 GPUs. Profiling and evaluation details are provided in Appendices~\ref{app:training} and~\ref{app:benchmark}, respectively.

\paragraph{Baselines.}
We compare with NavDP~\citep{navdp2026}, ApexNav~\citep{apexnav2025}, StreamVLN~\citep{wei2026streamvln}, and Uni-NaVid~\citep{uninavid2025}. To adapt NavDP and ApexNav to our language-conditioned task, we use LocateAnything~\citep{locateanything2026} to ground instruction-specified objects in RGB observations. For NavDP, we back-project the initial detection using simulator-provided ground-truth depth to obtain a fixed world-frame goal. We express this goal as a 2D vector in the robot's current local frame at each step. For ApexNav, LocateAnything replaces the detectors in its category-level query pipeline. Both methods receive ground-truth depth with maximum ranges of 5~m for NavDP and 4.99~m for ApexNav, following their original evaluation settings.

For StreamVLN and Uni-NaVid, we evaluate both released checkpoints and versions fine-tuned on our action-labeled simulated data. Fine-tuning uses no additional DAgger~\citep{ross2011reduction} rollouts or reinforcement learning. Adaptation and fine-tuning details are provided in Appendix~\ref{app:benchmark}.

We also evaluate a variant without additional training on our Internet egocentric videos. This variant skips Stage 1 and starts directly from the pretrained Wan weights, retaining Stages 2 and 3 on action-labeled simulated data but omitting video replay.

\label{sec:exp-main}
\begin{table}[t]
\caption{Comparison on Short-Horizon OVON and HabitatGS-LN.
``Internet videos'' refers to our curated Internet egocentric videos. Best results are in \textbf{bold}.
Dashes indicate model variants not profiled separately for latency.
$\dagger$ We evaluate a single-view subset of Short-Horizon OVON under our evaluation protocol.}
\label{tab:main}
\centering
\begingroup
\setlength{\tabcolsep}{2.5pt}
\renewcommand{\arraystretch}{1.1}
\begin{tabular*}{\linewidth}{@{\extracolsep{\fill}}l*{4}{>{\raggedleft\arraybackslash}p{0.64in}}r@{}}
\toprule
\multirow{2}{*}{Method}
& \multicolumn{2}{c}{Short-Horizon OVON$^\dagger$}
& \multicolumn{2}{c}{HabitatGS-LN}
& \multirow{2}{*}{\shortstack{Amortized\\latency (ms) $\downarrow$}} \\
\cmidrule(lr){2-3}\cmidrule(lr){4-5}
& SR (\%) $\uparrow$ & SPL (\%) $\uparrow$ & SR (\%) $\uparrow$ & SPL (\%) $\uparrow$ & \\
\midrule
LocateAnything + NavDP   & 74.4 & 37.2 & 82.8 & 61.0 & 184 \\
LocateAnything + ApexNav & 73.3 & 64.1 & 75.8 & 66.8 & 511 \\
StreamVLN (zero-shot)    & 61.1 & 51.2 & 48.4 & 40.0 & 87 \\
Uni-NaVid (zero-shot)    & 11.6 & 9.9 & 8.9 & 8.2 & 99 \\
\midrule
StreamVLN (fine-tuned)   & 79.6 & 62.9 & 75.8 & 59.5 & -- \\
Uni-NaVid (fine-tuned)   & 80.2 & 68.1 & 82.2 & 71.6 & -- \\
\midrule
Ours (w/o Internet videos) & 81.1 & 65.5 & 90.4 & 80.8 & -- \\
\textbf{Ours} & \textbf{86.5} & \textbf{71.3} & \textbf{94.9} & \textbf{86.1} & \textbf{76} \\
\bottomrule
\end{tabular*}
\endgroup
\end{table}

\paragraph{Comparative results.}
As shown in Table~\ref{tab:main}, ScaleNav outperforms the evaluated baselines in SR and SPL on both benchmarks. Compared with our variant without Internet egocentric videos, SR improves by 5.3 and 4.5 percentage points on Short-Horizon OVON and HabitatGS-LN, respectively. These gains support the benefit of learning navigation priors from Internet egocentric videos. Our framework also achieves a comparatively low amortized inference latency of 76~ms per executed action.

\subsection{Ablation Studies}
\label{sec:exp-ablation}
We conduct the following ablation studies on HabitatGS-LN.

\begin{table}[!htbp]
\caption{Internet egocentric video utilization and component ablations on HabitatGS-LN.
Video pretraining and Alignment denote Stages 1 and 2, respectively.
Pseudo-action denotes action labels estimated from these videos,
Anchor denotes the initial-frame anchor, and Replay denotes
Internet egocentric video replay during Stages 2 and 3. Best results are highlighted in \textbf{bold}.}
\label{tab:pseudolabel-mix}
\centering
\setlength{\tabcolsep}{3pt}
\begin{tabular}{@{}lccccc cc@{}}
\toprule
Variant & \shortstack{Video\\pretraining} & \shortstack{Pseudo-\\action} & Anchor & Alignment & Replay & SR (\%) $\uparrow$ & SPL (\%) $\uparrow$ \\
\midrule
w/o Internet videos      & -- & -- & $\checkmark$ & $\checkmark$ & -- & 90.4 & 80.8 \\
Pseudo-action training   & -- & $\checkmark$ & $\checkmark$ & $\checkmark$ & -- & 73.9 & 67.1 \\
w/o video pretraining    & -- & -- & $\checkmark$ & $\checkmark$ & $\checkmark$ & 90.4 & 83.1 \\
\midrule
w/o initial-frame anchor & $\checkmark$ & -- & -- & $\checkmark$ & $\checkmark$ & 89.8 & 83.1 \\
w/o action alignment     & $\checkmark$ & -- & $\checkmark$ & -- & $\checkmark$ & 93.6 & 85.1 \\
w/o video replay         & $\checkmark$ & -- & $\checkmark$ & $\checkmark$ & -- & 94.3 & 84.2 \\
\midrule
\textbf{Full model}      & $\checkmark$ & -- & $\checkmark$ & $\checkmark$ & $\checkmark$ & \textbf{94.9} & \textbf{86.1} \\
\bottomrule
\end{tabular}
\end{table}

\paragraph{Internet egocentric video utilization.}
We compare different ways of learning from Internet egocentric videos in Table~\ref{tab:pseudolabel-mix}. We use Depth Anything 3~\citep{lin2025depthanything3} to estimate camera poses and derive pseudo action labels. The resulting video-action pairs are mixed with action-labeled simulated data during Stages 2 and 3, with equal sampling probability per clip and no Stage 1 pretraining. This yields 73.9\% SR and 67.1\% SPL, below the variant without Internet egocentric videos. Pose estimation errors may contribute to this degradation by introducing noise into action supervision.

We also consider jointly training on Internet egocentric videos and action-labeled simulated data during Stages 2 and 3, without a separate video pretraining stage. The Internet videos remain action-free and contribute only the video loss. This approach achieves 90.4\% SR and 83.1\% SPL, compared with 94.9\% and 86.1\% for our full model. These results suggest that dedicated video pretraining is important for learning navigation priors from Internet egocentric videos to support subsequent action generation.

\paragraph{Component ablations.}
Table~\ref{tab:pseudolabel-mix} shows that removing the initial-frame anchor while retaining recent history reduces SR from 94.9\% to 89.8\%, supporting the benefit of preserving the initial visual reference. Skipping Stage 2 or omitting Internet egocentric video replay during Stages 2 and 3 yields 93.6\% and 94.3\% SR, respectively, suggesting modest benefits from action alignment and video replay.

\subsection{Real-Robot Deployment}
\label{sec:exp-robot}
We further evaluate ScaleNav on a Unitree Go2 in an exhibition hall, an open lounge area, and an office area. Each method is evaluated on four tasks per environment with five trials per task, totaling 60 trials. We compare ScaleNav with its variant without Internet egocentric videos, StreamVLN, and Uni-NaVid, using the same initial poses and instructions.

We manually assess success using the stopping and footprint-distance criteria in Section~\ref{sec:exp-setup}. Further experimental details are provided in Appendix~\ref{app:robot}.

\begin{table}[!htbp]
\caption{Real-robot success rates (60 trials per method). Best results are in \textbf{bold}.}
\label{tab:real-robot}
\centering
\begingroup
\renewcommand{\arraystretch}{1.1}
\begin{tabular*}{\linewidth}{@{\extracolsep{\fill}}lrrrr@{}}
\toprule
\multirow{2}{*}{Method} & \multicolumn{4}{c}{SR (\%) $\uparrow$} \\
\cmidrule(lr){2-5}
& Hall & Lounge & Office & Overall \\
\midrule
StreamVLN (fine-tuned) & 25.0 & 60.0 & 40.0 & 41.7 \\
Uni-NaVid (fine-tuned) & 60.0 & 75.0 & 20.0 & 51.7 \\
\midrule
Ours (w/o Internet videos) & 75.0 & 60.0 & 60.0 & 65.0 \\
\textbf{Ours} & \textbf{85.0} & \textbf{85.0} & \textbf{80.0} & \textbf{83.3} \\
\bottomrule
\end{tabular*}
\endgroup
\end{table}

\begin{figure}[t]
\centering
\includegraphics[width=\linewidth]{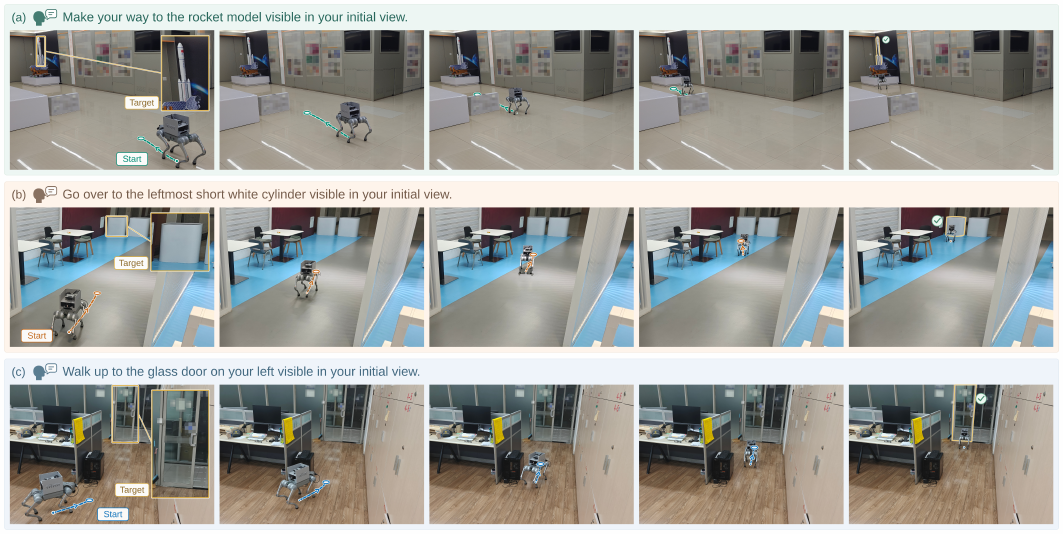}
\caption{\textbf{Qualitative results of ScaleNav} in an exhibition hall (a), an open lounge area (b), and an office area (c). The robot follows language instructions to reach the specified targets.}
\label{fig:real-robot-qualitative}
\end{figure}

ScaleNav achieves 83.3\% overall SR and the highest SR among the evaluated methods in all three environments (Table~\ref{tab:real-robot}). These results support its ability to generalize to real-world navigation. Figure~\ref{fig:real-robot-qualitative} shows qualitative results.

\section{Conclusion}
\label{sec:conclusion}
We presented ScaleNav, a language-conditioned local navigation framework that learns navigation priors through Internet egocentric video pretraining and translates them into executable action chunks through action alignment and joint fine-tuning. Simulation experiments demonstrate the benefits of Internet egocentric videos and improvements over the evaluated baselines in success rate and path efficiency. Real-robot experiments further support generalization to real-world navigation. For limitations, our quantitative evaluation focuses on indoor navigation to initially visible object targets. Generalization to outdoor environments and navigation to region-level goals remain to be evaluated.

\subsection*{AI use statement}
In preparing this manuscript, we used generative AI tools to improve readability and organize the presentation. We also used a vision-language model to generate navigation instructions and assist with data screening, as described in Section~\ref{sec:data}. The authors reviewed and verified the AI-assisted manuscript text and take responsibility for the final submission.

\setlength{\bibsep}{4pt plus 1pt minus 1pt}
\bibliography{references}
\bibliographystyle{iclr2027_conference}

\clearpage
\appendix
\section{Qualitative Analysis of Video Pretraining}
\label{sec:exp-video-vis}
\begin{figure}[htbp]
\centering
\includegraphics[width=\linewidth]{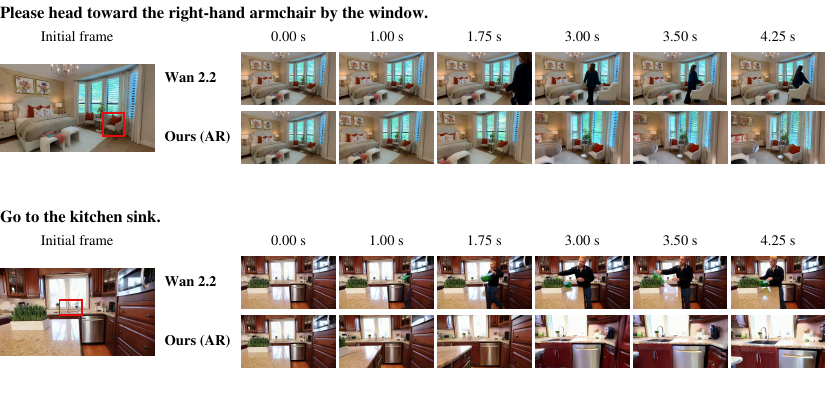}
\caption{Qualitative comparison of Wan 2.2 and our Stage 1 model with autoregressive (AR) video generation. Red boxes mark the navigation targets.}
\label{fig:video-pretraining-vis}
\end{figure}

Figure~\ref{fig:video-pretraining-vis} compares the original Wan2.2-TI2V-5B model with our model after Stage 1, given the same initial frame and navigation instruction. Wan 2.2 generates entire clips bidirectionally, while our model generates chunks autoregressively. In both examples, Wan 2.2 animates a human subject unrelated to navigation and keeps the camera largely stationary, whereas our model predicts camera motion toward the target.

\FloatBarrier
\section{Implementation Details}
\label{app:training}

\paragraph{Model and input configuration.}
We use RGB frames at a resolution of $512\times288$ pixels and a sampling rate of 4~Hz. Each video chunk contains nine observations, $o_i,\ldots,o_{i+8}$, with $o_i$ serving as the current-observation condition. Its $T=9$ actions correspond to the transitions $o_i\to o_{i+1},\ldots,o_{i+8}\to o_{i+9}$, where $o_{i+9}$ is the first observation of the next chunk. The VAE encodes the nine frames into three temporal latent slices. The history window retains the two most recent chunks in addition to the initial-frame anchor. Each action component $u$ is normalized using a scaled symlog transformation~\citep{hafner2025dreamerv3}, $f(u)=\operatorname{sign}(u)\log(1+|u|/c)$, with $c=0.01\,\mathrm{m}$ for translation and $c=0.01\,\mathrm{rad}$ for yaw. We recover action components using $f^{-1}(z)=c\,\operatorname{sign}(z)(\exp(|z|)-1)$.

\paragraph{Training configuration.}
We use AdamW~\citep{loshchilov2019adamw} with BF16 mixed precision throughout training. Table~\ref{tab:training-config} summarizes the stage-specific settings. Video pretraining consists of 2,000 updates under the original video-training formulation, followed by 8,000 updates with chunk-wise causal prediction. For joint fine-tuning, we warm up the learning rate over 500 updates, hold it constant until update 4,000, and apply cosine decay to $5\times10^{-7}$ by update 5,500. It remains constant thereafter, and we use the checkpoint at update 6,500 for evaluation.
\begin{table}[htbp]
\caption{Training configuration. Update counts refer to the checkpoints used in the final model's training sequence. Video replay probabilities specify dataset sampling, not loss weights. The dash indicates that Stage 1 uses only Internet egocentric videos, rather than a replay mixture.}
\label{tab:training-config}
\centering
\begin{tabular}{lccc}
\toprule
Configuration & Video pretraining & Action alignment & Joint fine-tuning \\
\midrule
Updates & 10,000 & 2,500 & 6,500 \\
Global batch size & 64 & 64 & 32 \\
Peak learning rate & $1\times10^{-5}$ & $1\times10^{-5}$ & $5\times10^{-6}$ \\
Video replay probability & -- & 10\% & 20\% \\
\bottomrule
\end{tabular}
\end{table}

\paragraph{Inference and execution.}
We jointly generate video latents and action chunks using four Euler denoising steps. The classifier-free guidance~\citep{ho2022cfg} scales are 5.0 for video and 1.0 for actions, with noise-schedule shift factors of 5.0 and 1.0, respectively. At each denoising step, the model receives the video and action noise levels from their respective schedules, and each modality is updated using its own Euler step size. The unconditional branch removes the text condition while retaining the current observation and history. During simulation evaluation, we execute the nine actions in each chunk sequentially at 4~Hz and predict the next chunk from the latest observation, unless the episode terminates. The visual history is constructed from observations collected during execution rather than generated frames. During simulation evaluation, action history stores the model's previous action predictions rather than measured robot displacements.

\paragraph{Inference optimizations.}
Historical tokens use fixed zero-noise timesteps and cannot attend to the current chunk. We therefore compute their keys and values once for each classifier-free guidance branch and reuse them across denoising steps~\citep{ye2026dreamzero}. We batch the conditional and unconditional forward passes for classifier-free guidance and use CUDA graphs\footnote{\url{https://docs.nvidia.com/cuda/cuda-programming-guide/04-special-topics/cuda-graphs.html}} to reduce kernel-launch overhead. We also omit RGB decoding of the predicted video latents because action execution does not require the generated frames. At four denoising steps, these optimizations reduce per-chunk policy-call latency from 1,791 to 684~ms, a $2.6\times$ speedup (Table~\ref{tab:latency}). The step-count sweep measures latency only, not navigation success.

We profile ScaleNav on recorded observations in BF16 using CUDA synchronization and the guidance settings above. Table~\ref{tab:latency} reports mean per-chunk policy-call latency, including current-observation encoding, completed-chunk history encoding, history KV prefill, denoising, and buffered-action retrieval. Video decoding is included only in the ablation configurations that retain it. Measurements exclude simulator rendering, network communication, and motion execution.

The four-step configurations use 23 measured chunks after five warm-up chunks per configuration. Configurations with more denoising steps use 10 measured chunks after four warm-up chunks. Over the 23 measured chunks, the fully optimized four-step implementation has a sample standard deviation of 0.5~ms and an empirical 95th percentile of 685.1~ms for per-chunk policy-call latency.

The baseline profiling scripts time policy calls, including preprocessing and model inference, as well as object grounding, segmentation, semantic fusion, map updates, and planning where applicable. Simulator rendering and motion execution are excluded.

\begin{table}[htbp]
\caption{Mean per-chunk policy-call latency on an RTX 4090 (ms), including history updates. \textbf{Left:} unoptimized and fully optimized implementations at different denoising step counts. \textbf{Right:} cumulative optimizations at four steps. Speedups are relative to the unoptimized implementation at the same step count.}
\label{tab:latency}
\centering
\setlength{\tabcolsep}{3pt}
\begin{tabular}{crrc}
\toprule
Steps & Unoptimized & Optimized & Speedup \\
\midrule
4  & 1791  & \textbf{684} & 2.6$\times$ \\
8  & 3272  & 1034 & 3.2$\times$ \\
20 & 7731  & 2080 & 3.7$\times$ \\
50 & 18846 & 4693 & 4.0$\times$ \\
\bottomrule
\end{tabular}
\hfill
\begin{tabular}{lrc}
\toprule
Optimization (cumulative) & Latency & Speedup \\
\midrule
Unoptimized & 1791 & 1.0$\times$ \\
+ History KV caching & 967 & 1.9$\times$ \\
+ Batched guidance & 948 & 1.9$\times$ \\
+ CUDA graphs & 920 & 1.9$\times$ \\
+ Omit video decoding & \textbf{684} & 2.6$\times$ \\
\bottomrule
\end{tabular}
\end{table}

\section{Data Construction Details}
\label{app:data}

We use Gemini 3.5 Flash\footnote{\url{https://ai.google.dev/gemini-api/docs/models/gemini-3.5-flash}} for VLM-based annotation and instruction paraphrasing. Figure~\ref{fig:data-examples} shows examples from the two training datasets.
\begingroup
\newcommand{\dataexampleframe}[5]{%
  \begin{minipage}[t]{.327\linewidth}
  \vspace{0pt}
  \setlength{\unitlength}{\dimexpr\linewidth/512\relax}
  \begin{picture}(512,288)
  \put(0,0){#1}
  \put(#2,#3){\color{red}\linethickness{.65pt}\framebox(#4,#5){}}
  \end{picture}
  \end{minipage}%
}
\begin{figure}[htb]
\centering
{\raggedright\small\textbf{Internet egocentric video}\par}\vspace{3pt}
\dataexampleframe{\includegraphics[width=\linewidth]{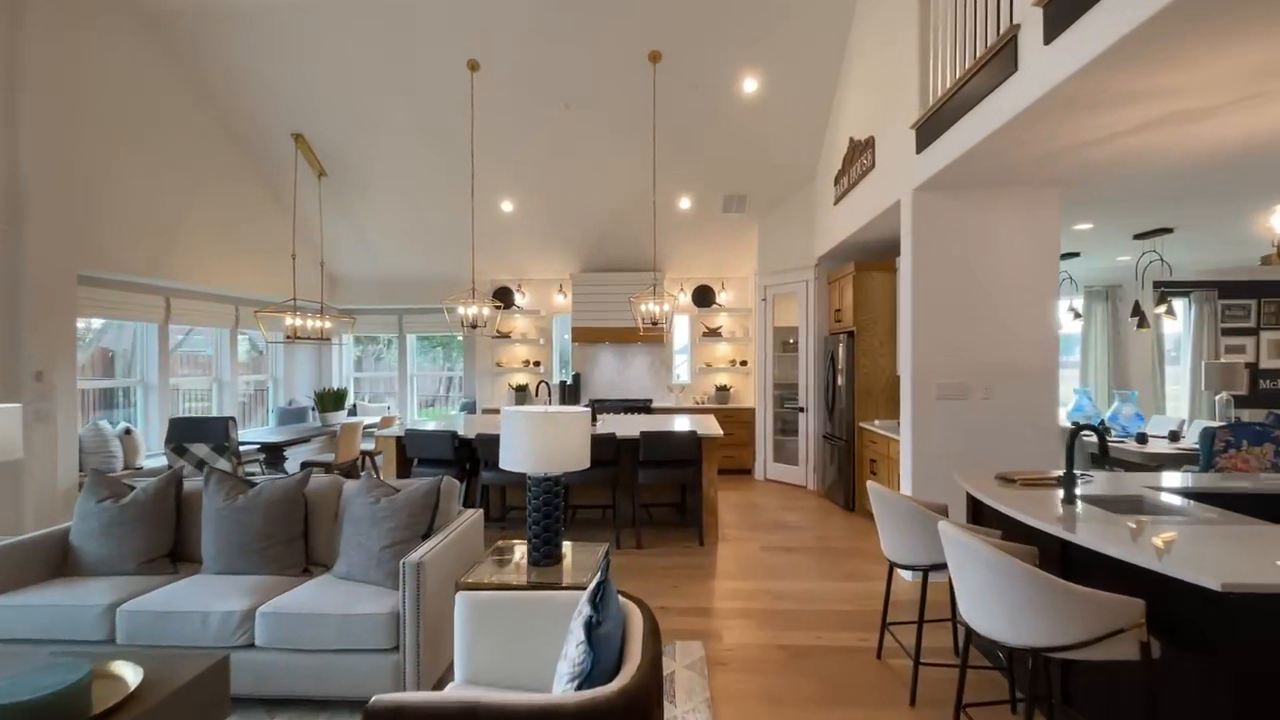}}{278}{95}{25}{32}\hfill
\dataexampleframe{\includegraphics[width=\linewidth]{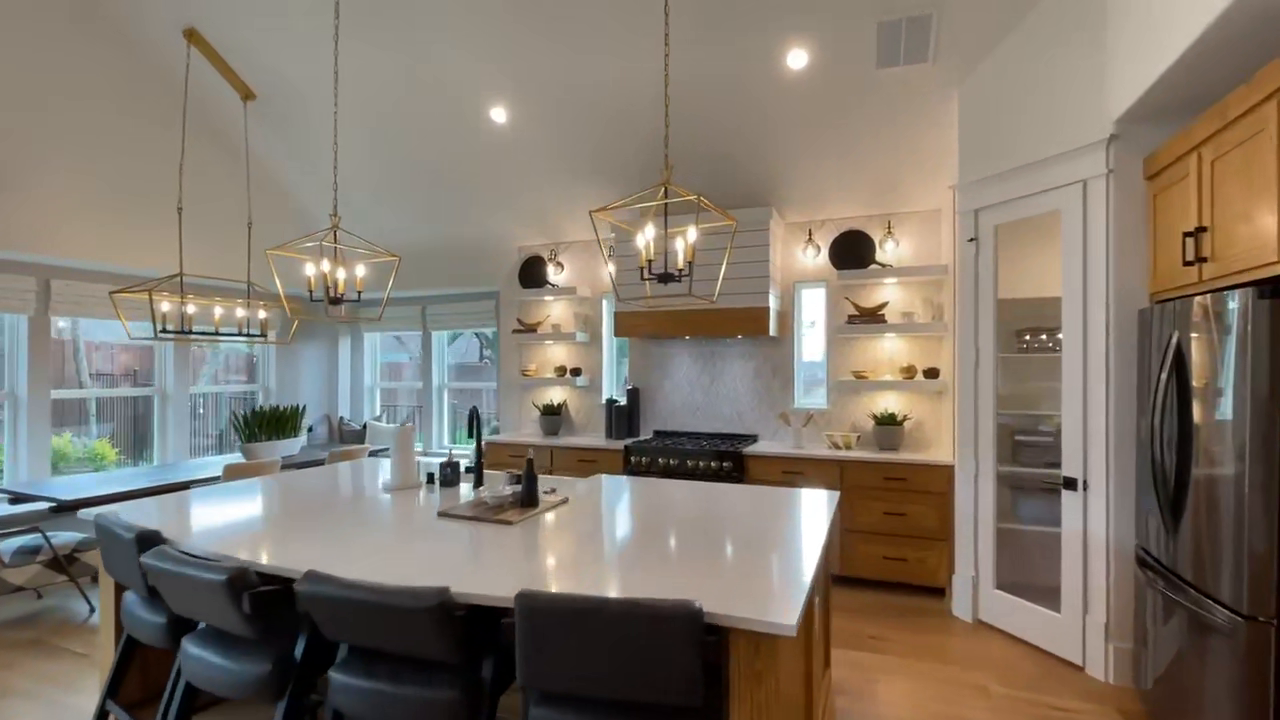}}{298}{50}{84}{58}\hfill
\dataexampleframe{\includegraphics[width=\linewidth]{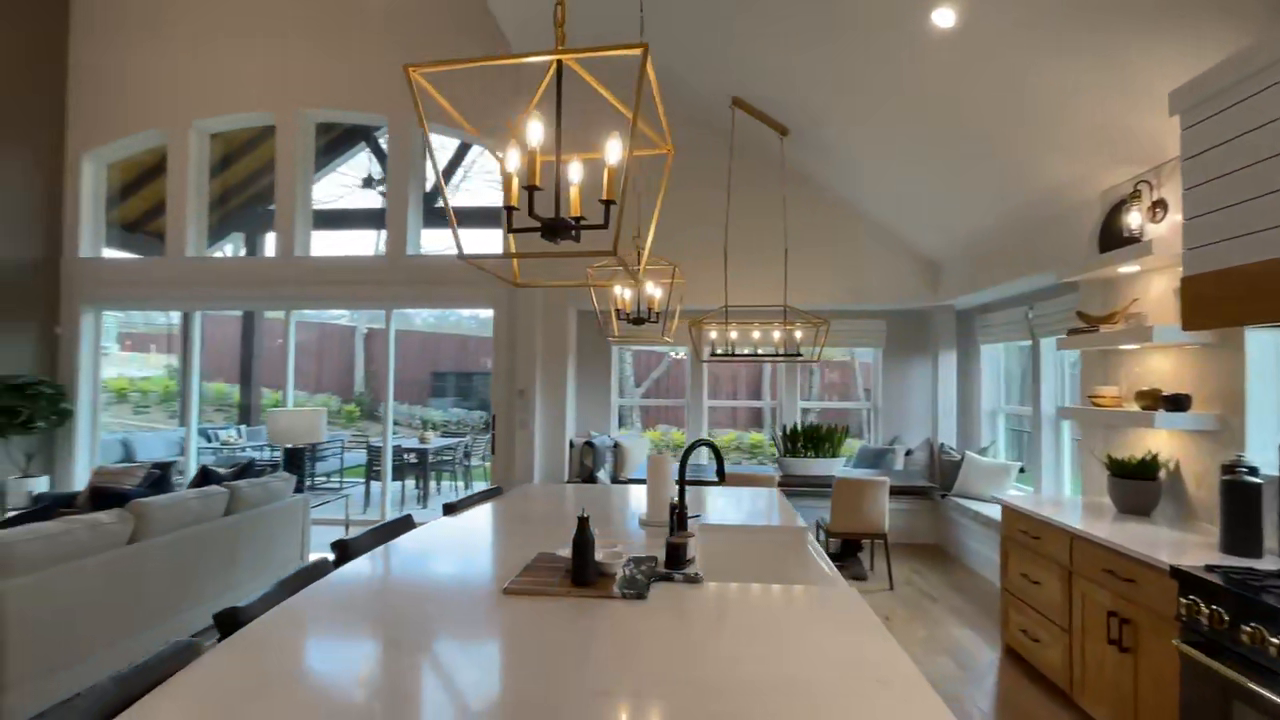}}{398}{1}{77}{89}
\par\vspace{3pt}
{\raggedright\footnotesize ``Go to the wooden kitchen cabinets.''\par}
\vspace{8pt}
{\raggedright\small\textbf{Action-labeled simulated data}\par}\vspace{3pt}
\dataexampleframe{\includegraphics[width=\linewidth]{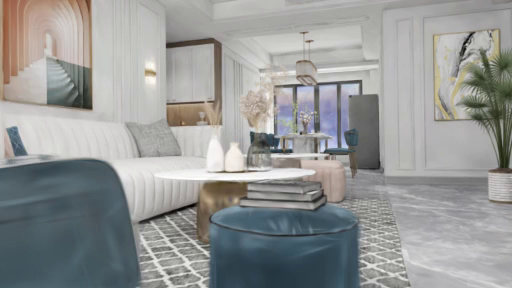}}{325}{111}{36}{51}\hfill
\dataexampleframe{\includegraphics[width=\linewidth]{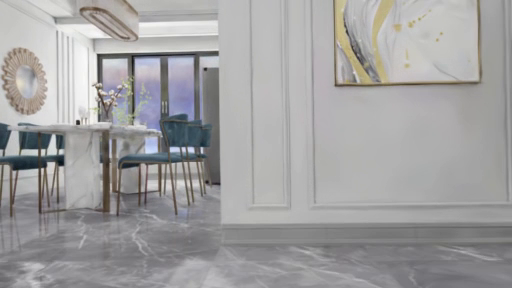}}{111}{68}{102}{102}\hfill
\dataexampleframe{\includegraphics[width=\linewidth]{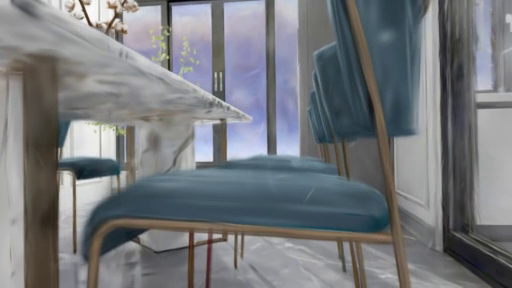}}{62}{1}{385}{286}
\par\vspace{3pt}
{\raggedright\footnotesize ``Navigate to the blue chair next to the refrigerator visible in your initial view.''\par}
\caption{Examples from the two training datasets, paired with their navigation instructions. Frames are shown in temporal order. Red boxes mark the targets for illustration only.}
\label{fig:data-examples}
\end{figure}
\endgroup

\subsection{Internet Egocentric Videos}

\paragraph{Segment extraction.}
For each source clip, we sample keyframes at uniform intervals and provide them in temporal order to the VLM. The extraction prompt takes keyframes labeled with source-frame indices and returns segment boundaries and instructions. We map the returned indices to timestamps in the source clip. Initial instructions follow the form ``go to \ldots{}'' and describe the selected target by its category, attributes, or spatial relations.
\begin{Verbatim}
Identify navigation segments from the supplied egocentric video keyframes. The keyframes are sampled at uniform intervals, presented in temporal order, and labeled with source-frame indices.

For each segment:
1. Identify a frame in which the camera has arrived near a clearly describable object or region. Treat this frame as a candidate endpoint and the corresponding object or region as the navigation target.
2. Search backward through the supplied keyframes and identify the earliest keyframe in which the same target is visible. Use this as the starting frame. The start must precede the endpoint.
3. Generate a concise navigation instruction in the form "Go to ...", describing the target by its category, attributes, or spatial relations. Ground the description in the initial observation.

Return JSON only, with a "candidates" list. Each entry must contain:
"target_description", "start_frame", "end_frame", and "instruction". Use only the supplied frame indices. Return {"candidates": []} if no valid segment is identified.
\end{Verbatim}

\paragraph{Segment review.}
The segment-review prompt below assesses target-instruction consistency, motion plausibility, arrival accuracy, and temporal continuity. We separately label the candidate segment and the surrounding endpoint context to support arrival assessment.
\begin{Verbatim}
Review the candidate navigation segment and its instruction.

Inputs:
- Instruction: {instruction}
- Target description: {target_description}
- Segment boundaries: {start_frame}, {end_frame}
- Chronologically ordered frames covering the segment.
- Separately labeled source frames around the candidate endpoint.

Assess the following four criteria:
1. Target-instruction consistency: Verify that the initial and final frames depict the same target and that its category and visible attributes match the instruction. Check that any stated spatial relations are supported by the initial observation, using that view as the reference for left/right and ordinal descriptions. When multiple plausible targets are visible, verify that the instruction uniquely identifies the intended target. Mark this criterion as "fail" if the wording is ambiguous, or "uncertain" if the visual evidence is insufficient to decide.
2. Motion plausibility: Check that the segment primarily shows camera movement toward the target, rather than prolonged in-place rotation, stationary observation, or back-and-forth motion that does not advance toward it.
3. Arrival accuracy: Examine the frames around the candidate endpoint to determine whether the camera has reached the target's vicinity.
4. Temporal continuity: Check for scene cuts or abrupt temporal jumps within the selected segment.

Return JSON with the keys "target_instruction_consistency", "motion_plausibility", "arrival_accuracy", and "temporal_continuity". For each criterion, provide a "decision" ("pass", "fail", or "uncertain") and a brief "reason" grounded in the supplied frames. Include "keep", set to true only when all four criteria pass.
\end{Verbatim}

\subsection{Action-Labeled Simulated Data}

\paragraph{Scene and target selection.}
Our candidate pool contains approximately 1K \href{https://huggingface.co/datasets/spatialverse/InteriorGS}{InteriorGS} scenes~\citep{miao2026towards}, each with instance-level semantic 3D bounding boxes and room layout annotations. We exclude scenes with poor rendering quality or inadequate semantic annotations. We randomly sample starting poses from navmesh-defined traversable areas. Given the initial observation and scene annotations, we select a target and its corresponding goal position. We then form an initial instruction from the target's category or spatial relation and refine ambiguous references with the prompt below.

\paragraph{Trajectory generation and action labels.}
Our offline 2D adaptation of SCAN-Planner~\citep{zheng2026scanplanner} uses A$^*$~\citep{hart1968astar} to guide local obstacle avoidance and optimizes the path while accounting for the robot's footprint. The planner then performs time reparameterization and collision checking. We sample egocentric observations and robot poses at 4~Hz while executing the time-parameterized trajectory. We convert consecutive global poses into relative poses in the preceding robot body frame to obtain action labels.

\paragraph{Instruction refinement.}
The following prompt refines ambiguous target references without changing the selected target or destination.
\begin{Verbatim}
Refine the navigation instruction to distinguish the selected target from other objects of the same category visible in the initial observation.

Inputs:
- Initial egocentric observation.
- Initial instruction: {instruction}
- Selected target: {selected_target}
- Spatial and semantic information for the relevant objects:
  {object_annotations}

Preserve the selected target and intended destination. When the category name alone is ambiguous, add a concise description of the target's visible attributes or relative position, such as "the red chair on the right". Interpret directional descriptions from the initial observation.

Use the scene annotations to associate the selected target with the image. Include only attributes and spatial relations supported by the initial observation. Do not change the target, invent distinguishing features, or introduce additional navigation requirements.

If the original instruction already identifies the intended target unambiguously, retain it.

Return JSON with:
- "instruction": the refined instruction, or null if the target cannot be distinguished using the supplied information.
- "reason": a brief explanation of the distinguishing visual evidence.
\end{Verbatim}

\paragraph{Review with scene annotations.}
The following addendum supplements the segment-review prompt with scene annotations and a target-marked reference image.
\begin{Verbatim}
Apply the segment-review criteria to the supplied simulated clip. For arrival accuracy, use the final observation and nearby frames from the rendered sequence.

Additional inputs:
- Scene semantic annotations: {scene_annotations}
- A target-marked reference image identifying the intended target.

When assessing target-instruction consistency, combine these inputs with the unmarked observations to verify that the instruction refers to the selected target. Check that its category, distinguishing attributes, and stated spatial relations are supported by the initial observation. The reference marker identifies the intended target and is not a property of the object or part of the navigation instruction.

Use the same four assessment keys and output format as the segment-review template. Include the additional evidence in the reason for "target_instruction_consistency".
\end{Verbatim}

\subsection{Instruction Paraphrasing}

We use the following prompt to paraphrase accepted instructions from both datasets.
\begin{Verbatim}
Paraphrase the accepted navigation instruction to increase language diversity while preserving its meaning.

Original instruction: {instruction}

Preserve the intended target, destination, distinguishing attributes, and spatial relations. Preserve the directional references in the original instruction. Do not introduce new navigation requirements or remove information needed to identify the intended target.

Return JSON only, with the paraphrased instruction under "instruction".
\end{Verbatim}

\section{Benchmark and Evaluation Details}
\label{app:benchmark}

\paragraph{Single-view adaptation of Short-Horizon OVON.}
To adapt Short-Horizon OVON~\citep{gong2026abotn1} to our single-view setting, we retain episodes with an estimated target visibility of at least 50\% in the initial front view. We estimate visibility as the fraction of sampled vertices from the target's semantic mesh~\citep{yadav2023hm3dsem} whose projections fall within the same instance's region in the initial front-view semantic image. This is a target-visibility fraction, not the fraction of image pixels occupied by the target. The resulting evaluation set contains 524 episodes across 36 HM3D scenes~\citep{ramakrishnan2021hm3d}, each paired with a navigation instruction referring to the target object. Because this evaluation uses our single-view selection and the shared protocol in Section~\ref{sec:exp-setup}, its results are not directly comparable to results under the original three-view protocol.

\paragraph{HabitatGS-LN construction.}
We use indoor scenes from InteriorGS~\citep{miao2026towards} that are held out from training. Their scene IDs are disjoint from those of the 650 scenes used for our 2,600 action-labeled simulated training episodes. Using the simulation pipeline described in Section~\ref{sec:data}, we generate candidate episodes covering diverse object categories and spatial descriptions. Each episode specifies an initial robot pose and a navigation instruction that unambiguously refers to a specific object visible in the initial observation, rather than an object category. Although our task formulation allows both object and region targets, we restrict this benchmark to object targets for comparison with baselines designed for object-goal navigation.

\paragraph{Human review.}
To ensure the quality and validity of the evaluation tasks, two human reviewers independently inspect each candidate episode. They assess instruction accuracy and clarity, target visibility in the initial observation, destination reachability, and scene rendering quality. We retain only episodes approved by both reviewers, yielding 157 episodes across 45 scenes.

\begin{figure}[htbp]
\centering
\begin{minipage}[t]{.49\linewidth}
\vspace{0pt}
\setlength{\unitlength}{\dimexpr\linewidth/512\relax}
\begin{picture}(512,288)
\put(0,0){\includegraphics[width=\linewidth]{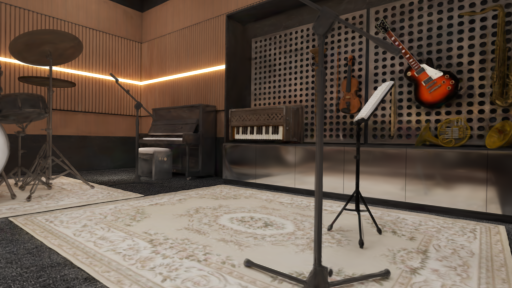}}
\put(97,99){\color{red}\linethickness{.65pt}\framebox(73,120){}}
\end{picture}\par\vspace{3pt}
{\raggedright\footnotesize ``please move to the microphone on your front-left visible in your initial view''\par}
\end{minipage}\hfill
\begin{minipage}[t]{.49\linewidth}
\vspace{0pt}
\setlength{\unitlength}{\dimexpr\linewidth/512\relax}
\begin{picture}(512,288)
\put(0,0){\includegraphics[width=\linewidth]{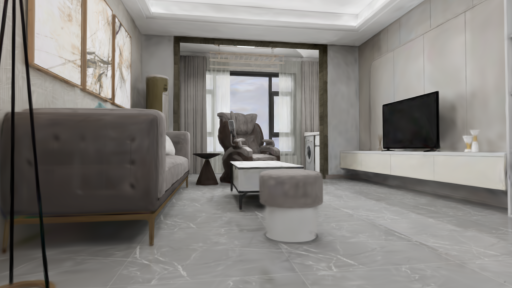}}
\put(201,98){\color{red}\linethickness{.65pt}\framebox(101,88){}}
\end{picture}\par\vspace{3pt}
{\raggedright\footnotesize ``Head to the massage chair ahead visible in your initial view.''\par}
\end{minipage}
\caption{Example tasks in HabitatGS-LN with their final navigation instructions. Red boxes mark the specified targets.}
\label{fig:benchmark-tasks}
\end{figure}

Figure~\ref{fig:benchmark-tasks} shows example tasks from HabitatGS-LN.

\paragraph{Simulation camera configuration.}
For all methods on both simulation benchmarks, we render RGB observations at a resolution of $512\times288$ pixels with a horizontal field of view of approximately $70^{\circ}$.

\paragraph{Simulation action execution.}
In both environments, we conduct closed-loop evaluation by executing the predicted actions and updating observations as the robot moves. We convert each method's discrete or continuous action outputs into kinematic pose updates. Within each scene, all methods use the same navmesh-based motion constraints.

\paragraph{NavDP goal construction.}
We compute the median of the valid depth values within the detected bounding box and select the pixel whose depth is closest to this median. Using its depth, the camera intrinsics, and the simulator-provided camera pose, we back-project this pixel into world coordinates without snapping the resulting point to the navmesh. We keep this world-frame goal fixed and express it as a 2D vector in the robot's current local frame at each step. NavDP predicts motion trajectories conditioned on this local goal and RGB-D observations.

\paragraph{ApexNav adaptation.}
ApexNav's original detection pipeline uses category-level queries. We replace YOLOv7~\citep{wang2023yolov7} and Grounding DINO~\citep{liu2024groundingdino} with LocateAnything to process the referring expressions in our instructions. The resulting detections feed into its segmentation, semantic fusion, and navigation pipeline.

\paragraph{Baseline fine-tuning.}
We prepare fine-tuning data for StreamVLN and Uni-NaVid from our action-labeled simulated training episodes. A path follower converts the reference trajectories into each model's native discrete actions. We execute these actions in Habitat-GS and re-render observations at the corresponding poses to ensure alignment between the visual observations and discrete action labels. We confirmed that all trajectories used for fine-tuning terminated near their intended goals. The resulting observation-action sequences are then paired with the navigation instructions and formatted for each model.

We adopt the optimization hyperparameters from StreamVLN's Stage 1 training script and Uni-NaVid's Stage 2 fine-tuning script. We initialize both models from their released checkpoints and fine-tune each model for five epochs using AdamW with a global batch size of 128. StreamVLN uses a base learning rate of $2\times10^{-5}$ and a learning rate of $5\times10^{-6}$ for its vision encoder. Uni-NaVid uses a learning rate of $10^{-5}$ and keeps its vision encoder frozen.

\paragraph{Footprint distance.}
The robot and target footprints are the regions they occupy when projected onto the ground plane. We model the robot footprint as an oriented capsule with a radius of 0.25~m and a centerline length of 0.36~m, using SCAN-Planner's Go2 parameters~\citep{zheng2026scanplanner}. We measure the minimum Euclidean distance between this capsule and the target's annotated ground-plane polygon~\citep{ericson2005realtime}.

\paragraph{SPL reference lengths.}
We compute reference lengths as the shortest navmesh distances to annotated goal positions on HabitatGS-LN and to sampled navigable points near valid target footprints on Short-Horizon OVON.

\paragraph{Stopping rules.}
Our model stops when five consecutive actions have translation magnitudes below 0.015~m and absolute yaw changes below 0.015~rad. NavDP stops when the robot is within 1~m of its estimated goal point on the ground plane. For the remaining baselines, we retain their default stopping mechanisms: explicit STOP actions for StreamVLN and Uni-NaVid, and rule-based stopping for ApexNav.

\paragraph{Pseudo-action labeling ablation.}
We use Depth Anything 3 (DA3-Nested)~\citep{lin2025depthanything3} to estimate metric camera poses from Internet egocentric videos. We invert the predicted world-to-camera transforms to obtain camera-to-world poses. We then express consecutive relative motions in a camera-aligned planar frame to obtain pseudo action labels. We retain clips with available pose estimates and do not apply confidence-based loss weighting.

\clearpage
\section{Real-Robot Evaluation}
\label{app:robot}

\paragraph{System.}
We deploy the policy on a Unitree Go2 equipped with a front-mounted Intel RealSense D435i camera and an onboard Jetson computer. The client sends rectified $512\times288$ RGB observations to a remote RTX 4090 GPU for inference. For both ScaleNav and the variant without Internet egocentric videos, we convert each predicted motion increment $a_t=(\Delta x_t,\Delta y_t,\Delta\psi_t)$ into body-frame linear and angular velocity references:
\[
v_t^x=\frac{\Delta x_t}{\Delta t},\qquad
v_t^y=\frac{\Delta y_t}{\Delta t},\qquad
\omega_t=\frac{\Delta\psi_t}{\Delta t},
\]
where $\Delta t=0.25~\mathrm{s}$ is the time interval represented by each action. These velocity references are passed to the robot's low-level motion interface for execution. After completing each action chunk, the robot holds still until the next prediction is available.

StreamVLN and Uni-NaVid predict discrete forward, left-turn, right-turn, and STOP actions. Forward actions specify a 0.25~m translation, while left/right turns specify $15^{\circ}$ rotations for StreamVLN and $30^{\circ}$ rotations for Uni-NaVid. All reported baseline trials use step-wise execution. The client uses robot odometry to measure progress and adjusts linear or angular velocity according to the remaining distance or yaw error. After completing an action, it acquires a new RGB observation before proceeding to the next action. A predicted STOP action terminates navigation.

\paragraph{Tasks and results.}
Figure~\ref{fig:realbot-tasks} presents the initial views and verbatim task instructions, and Table~\ref{tab:real-robot-tasks} reports per-task success counts. All methods use an execution budget of 200 actions per trial.

\begin{table}[!htbp]
\caption{Per-task real-robot results, reported as successes / trials. Best results in each row are in \textbf{bold}.}
\label{tab:real-robot-tasks}
\centering
\begingroup
\setlength{\tabcolsep}{4pt}
\renewcommand{\arraystretch}{1.0}
\begin{tabular*}{\linewidth}{@{\extracolsep{\fill}}lrrrr@{}}
\toprule
Task & \shortstack{StreamVLN\\(fine-tuned)} & \shortstack{Uni-NaVid\\(fine-tuned)} & \shortstack{Ours\\(w/o Internet videos)} & Ours \\
\midrule
\multicolumn{5}{@{}l}{\textit{Hall}} \\
Display counter & 2/5 & 4/5 & 4/5 & \textbf{5/5} \\
Rocket model & 1/5 & 1/5 & 2/5 & \textbf{3/5} \\
Screen on the rack & 0/5 & 4/5 & \textbf{5/5} & \textbf{5/5} \\
Second door & 2/5 & 3/5 & \textbf{4/5} & \textbf{4/5} \\
\midrule
\multicolumn{5}{@{}l}{\textit{Lounge}} \\
Wooden cabinet & 0/5 & 0/5 & 1/5 & \textbf{3/5} \\
Leftmost white cylinder & 3/5 & \textbf{5/5} & 4/5 & 4/5 \\
Second table & 4/5 & \textbf{5/5} & 3/5 & \textbf{5/5} \\
Vending machine & \textbf{5/5} & \textbf{5/5} & 4/5 & \textbf{5/5} \\
\midrule
\multicolumn{5}{@{}l}{\textit{Office}} \\
Cabinet under the printer & 0/5 & 0/5 & 1/5 & \textbf{3/5} \\
Left trash can & 0/5 & 4/5 & \textbf{5/5} & \textbf{5/5} \\
Glass door & 4/5 & 0/5 & 4/5 & \textbf{5/5} \\
Tripod phone stand & \textbf{4/5} & 0/5 & 2/5 & 3/5 \\
\bottomrule
\end{tabular*}
\endgroup
\end{table}

\begingroup
\newcommand{\realbottask}[7]{%
  \begin{minipage}[t]{0.24\linewidth}
  \vspace{0pt}\footnotesize
  \parbox[t][2\baselineskip][t]{\linewidth}{\raggedright\textbf{#1}}\par\vspace{3pt}
  \setlength{\unitlength}{\dimexpr\linewidth/512\relax}%
  \begin{picture}(512,288)
    \if\relax\detokenize{#2}\relax
      \put(0,0){\color{gray}\linethickness{0.3pt}\framebox(512,288){\shortstack{Correct initial view\\pending}}}%
    \else
      \put(0,0){#2}%
    \fi
    \ifnum#5>0\relax
      \put(#3,#4){\color{red}\linethickness{0.6pt}\framebox(#5,#6){}}%
    \fi
  \end{picture}\par\vspace{3pt}
  \parbox[t][6\baselineskip][t]{\linewidth}{\raggedright``#7''}%
  \end{minipage}%
}

\begin{figure}[!p]
\centering
{\raggedright\small\textbf{Hall}\par}\vspace{4pt}
\realbottask{(a) Display counter}
  {\includegraphics[width=\linewidth]{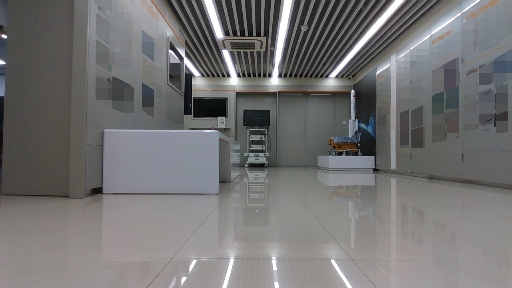}}{100}{91}{133}{69}
  {Head to the large white display counter on your left visible in your initial view.}
\hfill
\realbottask{(b) Rocket model}
  {\includegraphics[width=\linewidth]{figures/real_robot/redacted/hall_S1_main_20260923.png}}{345}{146}{23}{64}
  {Make your way to the rocket model visible in your initial view.}
\hfill
\realbottask{(c) Screen on the rack}
  {\includegraphics[width=\linewidth]{figures/real_robot/redacted/hall_S1_main_20260923.png}}{242}{158}{30}{24}
  {Walk over to the freestanding screen on the equipment rack visible in your initial view.}
\hfill
\realbottask{(d) Second door}
  {\includegraphics[width=\linewidth]{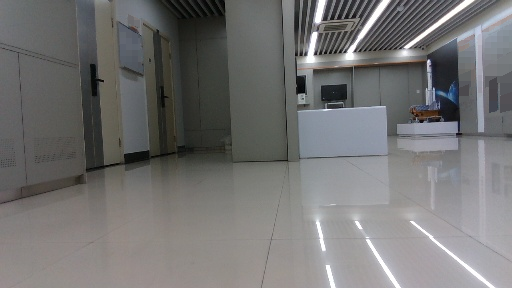}}{145}{124}{29}{130}
  {Walk over to the second door on your left visible in your initial view.}
\par\vspace{10pt}

{\raggedright\small\textbf{Lounge}\par}\vspace{4pt}
\realbottask{(e) Wooden cabinet}
  {\includegraphics[width=\linewidth]{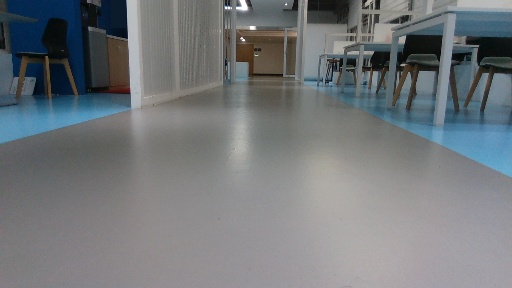}}{104}{196}{28}{58}
  {Walk over to the wooden cabinet on your left visible in your initial view.}
\hfill
\realbottask{(f) Leftmost white cylinder}
  {\includegraphics[width=\linewidth]{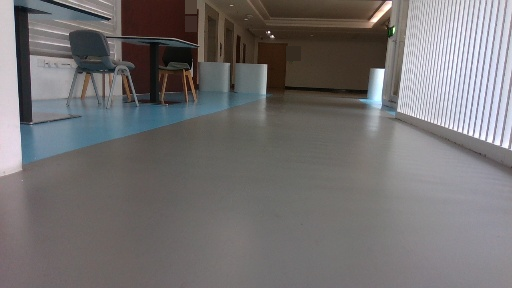}}{196}{190}{35}{39}
  {Go over to the leftmost short white cylinder visible in your initial view.}
\hfill
\realbottask{(g) Second table}
  {\includegraphics[width=\linewidth]{figures/real_robot/redacted/lounge_second_table_20260925_100732.png}}{64}{180}{139}{74}
  {Move toward the second table on your left visible in your initial view.}
\hfill
\realbottask{(h) Vending machine}
  {\includegraphics[width=\linewidth]{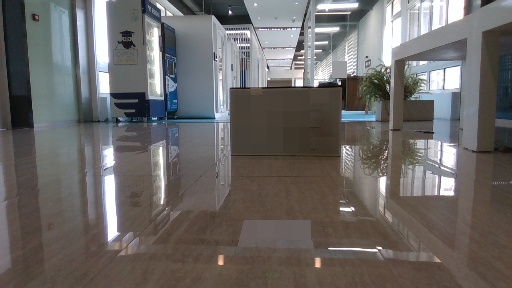}}{103}{162}{71}{126}
  {Proceed to the vending machine visible in your initial view.}
\par\vspace{10pt}

{\raggedright\small\textbf{Office}\par}\vspace{4pt}
\realbottask{(i) Cabinet under the printer}
  {\includegraphics[width=\linewidth]{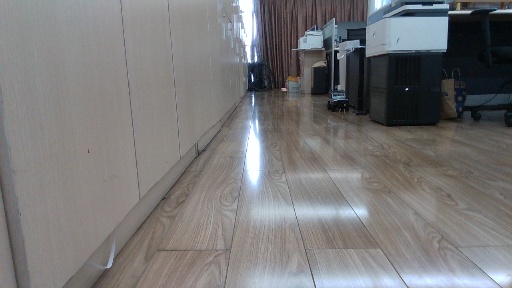}}{298}{190}{35}{50}
  {Go to the wooden cabinet under the printer visible in your initial view.}
\hfill
\realbottask{(j) Left trash can}
  {\includegraphics[width=\linewidth]{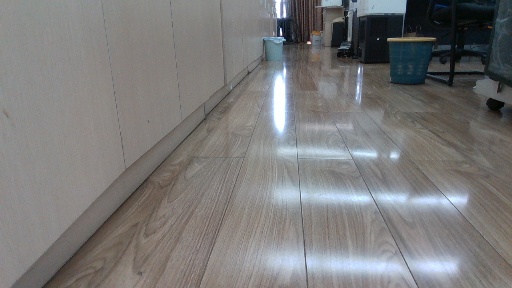}}{260}{225}{26}{28}
  {Reach the trash can on the left visible in your initial view.}
\hfill
\realbottask{(k) Glass door}
  {\includegraphics[width=\linewidth]{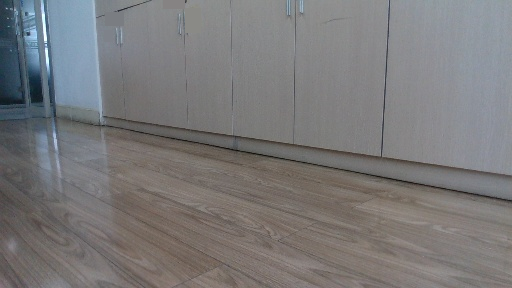}}{1}{168}{56}{119}
  {Walk up to the glass door on your left visible in your initial view.}
\hfill
\realbottask{(l) Tripod phone stand}
  {\includegraphics[width=\linewidth]{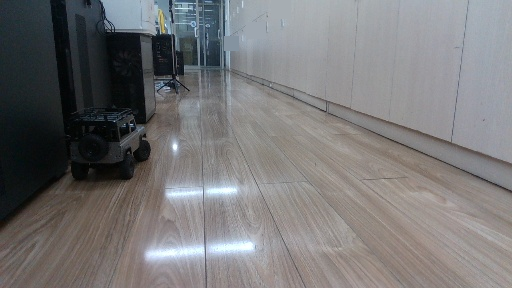}}{152}{191}{40}{97}
  {Approach the tripod phone stand visible in your initial view.}
\caption{Real-robot task views and instructions, grouped by environment. Red boxes mark the targets. Target boxes and privacy masks are for illustration only and are not policy inputs.}
\label{fig:realbot-tasks}
\end{figure}
\endgroup

\end{document}